\documentclass[letterpaper, 10 pt, conference]{ieeeconf}  

\IEEEoverridecommandlockouts                              

\usepackage{hyperref}
\usepackage{algorithm,algpseudocode}

\usepackage{amsmath}
\usepackage{amssymb}  

\usepackage{amsthm}
\usepackage{mathtools}
\usepackage{microtype}
\usepackage{graphicx}
\usepackage{subcaption}
\usepackage{booktabs} 
\usepackage{siunitx}

\theoremstyle{plain}
\newtheorem{theorem}{Theorem}[section]
\newtheorem{proposition}[theorem]{Proposition}

\theoremstyle{definition}
\newtheorem{definition}[theorem]{Definition}

\theoremstyle{remark}
\newtheorem{remark}[theorem]{Remark}

\usepackage{xcolor}

\usepackage{stfloats}

\usepackage[textsize=tiny]{todonotes}

\definecolor{Paired1}{RGB}{166,206,227}
\definecolor{Paired2}{RGB}{31,120,180}
\definecolor{Paired3}{RGB}{178,223,138}
\definecolor{Paired4}{RGB}{51,160,44}
\definecolor{Paired5}{RGB}{251,154,153}
\definecolor{Paired6}{RGB}{227,26,28}
\definecolor{Paired7}{RGB}{253,191,111}
\definecolor{Paired8}{RGB}{255,127,0}
\definecolor{Paired9}{RGB}{202,178,214}
\definecolor{Paired10}{RGB}{106,61,154}
\definecolor{Paired11}{RGB}{255,255,153}
\definecolor{Paired12}{RGB}{177,89,40}

\newcommand{\feature}{\ensuremath{\xi}}

\usepackage[backend=bibtex, natbib=true,style=ieee,
mincitenames=1,maxcitenames=2,maxbibnames=20,
url=false,doi=false]{biblatex}
\newif\ifanonymized
\anonymizedfalse

\title{
Observing and Controlling Features in Vision-Language-Action Models
}

\ifanonymized
  \author{Anonymized Authors}
\else
  \author{%
    Hugo Buurmeijer$^{1}$,
    Carmen Amo Alonso$^{1}$,
    Aiden Swann$^{2}$,
    and Marco Pavone$^{1,3}$%
    \thanks{$^{1}$Department of Aeronautics and Astronautics,
      Stanford University, USA. $^{2}$Department of Mechanical Engineering, Stanford University, USA.
    $^{3}$NVIDIA Research, USA.}%
  }
\fi

\begin{document}

\maketitle
\thispagestyle{empty}
\pagestyle{empty}

\begin{abstract}
Vision-Language-Action models (VLAs) have shown remarkable progress towards embodied intelligence.
While their architecture partially resembles that of Large Language Models (LLMs), VLAs exhibit higher complexity due to their multi-modal inputs/outputs and often hybrid nature of transformer and diffusion heads.
This is part of the reason why insights from mechanistic interpretability in LLMs, which explain how the internal model representations relate to their output behavior, do not trivially transfer to VLA counterparts.
In this work, we investigate whether VLA internal representations support lightweight behavioral steering without retraining.
Across four frontier VLA models, linear \emph{observers} recover state- and action-relevant information in both autoregressive and transformer--flow-matching architectures, and provide robust directions to causally alter VLA outputs.
Building on this, we introduce a \emph{controller} that minimally modifies representations to place observer predictions within prescribed target intervals.
Closed-loop experiments with $\pi_{0.5}$ in the LIBERO simulator and on DROID hardware demonstrate improved constraint satisfaction while retaining task performance, with only approximately $1\%$ inference overhead.
Together, these experiments show that lightweight linear interventions can reliably steer VLA behavior while preserving closed-loop capabilities, enabling alignment with user preferences and task requirements without fine-tuning.
\end{abstract}

\section{Introduction}

Vision-Language-Action models (VLAs) represent a significant step towards embodied intelligence.
By processing multi-modal inputs consisting of images, language instructions, and proprioceptive states, VLAs combine perception, reasoning, and action generation in a single model.
Similar to other disciplines where generative models have unlocked unprecedented generalization, VLAs enable robots to interpret natural language commands in rich visual contexts and execute complex behaviors, exhibiting improved generalization across tasks and environments \cite{intelligence2025pi05}.
However, VLAs suffer from similar limitations to other generative models: their behavior can be unpredictable, difficult to correct in real time, or misaligned with user preferences and safety requirements \cite{fei2025libero}.

One promising approach to overcoming these limitations is the ability to steer generative models toward desired behaviors while suppressing undesirable ones. In the context of Large Language Models (LLMs), intervening internal model representations to influence outputs has become a prominent research direction, referred to as ``activation steering"~\cite{turner2023activation}. Achieving a similar level of control is arguably even more critical for VLAs, where steering directly affects physical behavior in the real world. Furthermore, since VLAs often share architectural components with LLMs, (most notably transformer-based representations \cite{vaswani2017attention}), methods developed for LLMs could in theory transfer to VLAs without significant overhead. 
However, VLAs operate over multiple modalities, including proprioceptive inputs, and must function in closed-loop interaction with the physical world where compounding errors are notorious, thereby posing challenges absent in LLMs.
Additionally, robotic tasks such as manipulation naturally demand smooth, continuous actions, which has driven frontier architectures toward hybrid transformer-diffusion designs \cite{kawaharazuka2025vla-survey}, further complicating the transfer of interpretability insights from LLMs to VLAs.

To close this gap, several works have explored porting concepts from mechanistic interpretability in LLMs over to the domain of VLAs \cite{haon2025mechanistic,mitra2025mechanistic,lu2025probing,swann2026sparse}.
For instance, \cite{haon2025mechanistic} overwrites VLA model activations at inference time to modulate basic motion features such as speed and direction.
Overall, the effectiveness of existing methods is still limited. Crucially, while the focus of activation steering methods in LLMs is to control the behavior of the model while preserving its original merits, such as natural-sounding generations, in the field of VLAs interventions that preserve natural behaviors have not been studied.
Moreover, the effect of architectural design choices, such as purely transformer-based versus hybrid transformer-diffusion architectures, on the interpretability and steerability of VLA representations remains unexplored.

\begin{figure*}[t]
    \centering
    \includegraphics[width=0.95\textwidth]{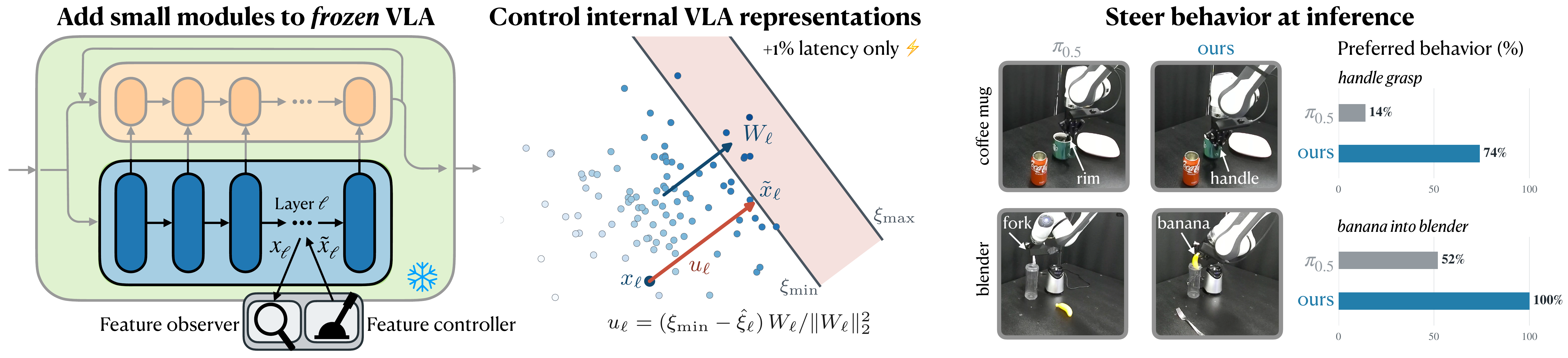}
    \vspace{-0.5em}
    \caption{We present the combination of \emph{feature-observability} and \emph{feature-controllability}: a framework to observe and control robot behavior through accessing and modifying internal VLA representations. A linear observer estimates features from the Transformer's internal representations, while a minimal additive controller steers these representations to align outputs with desired constraints. We demonstrate effective policy steering without fine-tuning in both simulation and on hardware.}
    \label{fig:hero_figure}
    \vspace{-1em}
\end{figure*}

In this paper, we address these challenges by formalizing concepts that are often implicit in work on steering LLMs, with the goal of making them applicable to VLAs.
We introduce a unified framework for analyzing and influencing VLA behavior through their internal representations (see Figure~\ref{fig:hero_figure}), and use them to systematically compare common VLA architectures.
We find linearly predictive state and action information in both autoregressive and transformer--flow-matching architectures, with performance varying across models, datasets, and features.
Building on these findings, we develop lightweight observers and representation controllers, and demonstrate behavioral steering while retaining task performance in closed-loop simulation and hardware experiments.

Concretely, our \textbf{contributions} are as follows:
\begin{itemize}
\item We formalize the concepts of \emph{feature-observability} and \emph{feature-controllability} for VLAs, inspired by classical notions of observability and controllability, providing a framework to analyze and influence behavior through internal representations. (Section \ref{sec:definitions})
\item We systematically evaluate linear observability of OpenVLA, $\pi_0$, $\pi_0$-FAST, and $\pi_{0.5}$ across LIBERO, DROID, and BridgeData V2 datasets.
We identify predictive information in two architecture families; states are generally more decodable than actions, with $\pi_{0.5}$ achieving the highest mean $R^2$ across continuous features on both LIBERO and DROID datasets. (Sections \ref{sec:feature-observer}, \ref{sec:feature_obs})
\item We propose a minimal linear controller that leverages linear observability, and show that steering along observed directions smoothly affects the model outputs. (Sections \ref{sec:feature-controller}, \ref{sec:feature_ctrl})
\item We present an online algorithm that integrates the observer and controller into closed-loop deployment without retraining.
Simulation and hardware experiments on $\pi_{0.5}$ demonstrate the effectiveness of the proposed representation intervention. (Sections \ref{sec:inference-time}, \ref{sec:closed_loop})
\end{itemize}

\section{Related Work}

\subsection{Mechanistic Interpretability in VLAs}

The intersection of mechanistic interpretability and robotics, specifically VLAs, remains significantly underexplored, though a number of recent works have investigated these tools in the context of end-to-end robot control.
Notably, steering vectors can be extracted from the feed-forward layers of the Transformer backbone to causally affect model behaviors such as speed and height~\cite{haon2025mechanistic}.
More abstract concepts, such as spatial relationships between objects and movement targets, were extracted from OpenVLA representations in~\cite{lu2025probing} using linear probes,
while~\cite{molinari2025emergent} probe for recoverable state-transition structure.
Such tools have also been used to improve downstream performance, e.g.\ by mechanistically identifying and fine-tuning task-relevant attention heads~\cite{mitra2025mechanistic}.

A complementary line of work forgoes pre-specified concepts in favor of unsupervised feature discovery: concurrent work of~\cite{swann2026sparse} trains Sparse Autoencoders on $\pi_{0.5}$ residual streams, proposes activation statistics to classify generality of features, and validates them via additive and ablative steering.
Our contributions are complementary: because we target features specified a priori, observability and controllability are measured against a known ground truth rather than against post-hoc inspection of top-activating episodes.

\subsection{Activation Steering in LLMs}
A growing body of work has shown that semantic and behavioral features in LLMs are often linearly represented in their activation spaces, a phenomenon commonly referred to as the \emph{linear separability hypothesis} \cite{park2024linear}. It has been widely observed that simple linear classifiers can reliably extract attributes such as sentiment, topic, or personality traits from intermediate transformer layers \cite{liseco,act,chen2025persona}, suggesting that these representations encode disentangled, behaviorally meaningful features. 
Building on this insight, \emph{activation steering} methods modify internal activations at inference time to influence model outputs without fine-tuning \cite{wu2024reft,sun2025hypersteer}. Techniques such as activation addition \cite{turner2023activation,li2023inference} have been applied to guide generation toward desired attributes, mitigate harmful behaviors, and improve safety and alignment. Crucially, these methods emphasize preserving the naturalness and coherence of generated text by applying low-rank or minimal perturbations to the model’s internal states.

\section{Problem Statement}\label{sec:problem-statement}

In this section, we present the problem of how the internal representations in VLAs can be controlled to align their outputs with user preferences. To that end, we describe the internal structure of two common VLA architectures, exemplified by OpenVLA \cite{OpenVLA} and $\pi_{0.5}$ \cite{intelligence2025pi05}. We then introduce two concepts that enable the formalization of the problem in mathematical terms.

\subsection{Vision-Language-Action Models (VLAs)}

\begin{figure}[t]
    \centering
    \includegraphics[width=\columnwidth]{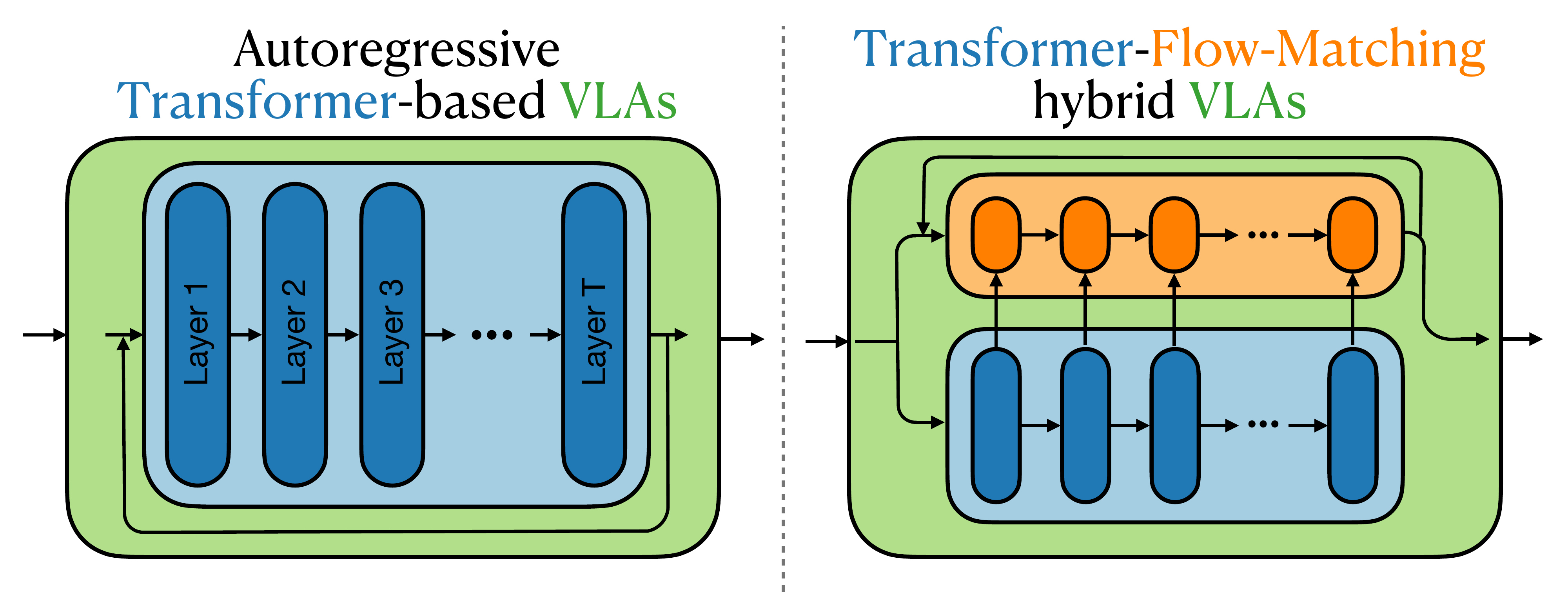}
    \vspace{-1em}
    \caption{Schematic representation of two \textcolor{Paired4}{VLA architectures}. On the left, a \emph{Transformer-based VLA} features an autoregressive \textcolor{Paired2}{Transformer architecture}. On the right, a \emph{Transformer-Flow-Matching hybrid VLA} is composed of a \textcolor{Paired2}{Transformer architecture} and a \textcolor{Paired8}{Flow Matching block} (referred to as the ``action expert"), where the Flow Matching layers attend to their paired counterparts in the Transformer.}
    \label{fig:architectures}
    \vspace{-1em}
\end{figure}

The goal is to formalize the study of internal representations in VLAs. Here we present two different types of \textcolor{Paired4}{VLA architectures} (Figure~\ref{fig:architectures}): (a) autoregressive \textcolor{Paired2}{transformer}-based, and (b) \textcolor{Paired2}{transformer}-\textcolor{Paired8}{flow-matching} hybrids.

\paragraph{\underline{\textbf{\textcolor{Paired2}{Transformer}}-based \textbf{\textcolor{Paired4}{VLAs}}}} This architecture consists of a Vision-Language-Model autoregressive transformer architecture, where visual observations (processed through vision encoders such as DINOv2 and SigLIP) and language instructions are encoded into a shared token space, and a language model backbone (e.g., Llama 2) predicts tokenized output actions autoregressively, which are then decoded into robot commands.
Examples of this architecture are OpenVLA \cite{OpenVLA}, $\pi_0$-FAST \cite{pertsch2025fast} and RT2 \cite{RT2}.

\paragraph{\underline{\textbf{\textcolor{Paired2}{Transformer}-\textcolor{Paired8}{Flow-Matching}} hybrid \textbf{\textcolor{Paired4}{VLAs}}}} This architecture consists of an autorregressive transformer architecture, where a pretrained Vision-Language Model (VLM) processes visual and language inputs, while a separate "action expert" transformer uses conditional flow matching to generate continuous, high-frequency action trajectories.
Examples of this architecture are $\pi_0$ \cite{blackPi0} and $\pi_{0.5}$ \cite{intelligence2025pi05}.

We formalize the \textcolor{Paired2}{Transformer architecture} \cite{vaswani2017attention} as a sequence of transformations (layers) acting on a latent representation of an input sequence. We denote the input by $w\in\mathcal{W}$, where $\mathcal{W}$ is the space of visual observations, language instructions, and proprioceptive states supplied to the model.
An encoding map $E$, incorporating modality-specific encoders and projections, maps $w$ to an initial hidden state $X_0=E(w)\in\mathbb{R}^{N\times d}$, where $N$ is the resulting number of token embeddings and $d$ is their dimension.
Then, each layer \( \mathsf{T}_\ell : \mathbb{R}^{N \times d} \to \mathbb{R}^{N \times d} \) transforms the hidden state from the previous layer, yielding
\begin{equation}\label{eqn:layers}
X_{\ell+1} = \mathsf{T}_{\ell+1}(X_{\ell}), \qquad \ell = 0, \dots, L-1,
\end{equation}
where $X_\ell \in \mathbb{R}^{N \times d}$ is the full hidden state after the $\ell$-th layer.

For the subsequent analysis, we summarize the full token matrix $X_\ell$ by its mean-pooled representation
\begin{equation}\label{eqn:mean_pooling}
x_{\ell}:=\frac{1}{N} \sum_{i=1}^N\left(X_{\ell}\right)_{i,:}^{\top} \in \mathbb{R}^d,
\end{equation}
where $\left(X_{\ell}\right)_{i,:}$ denotes the $i$-th row of $X_\ell$.
As is common for LLM representations, we assume mean pooling preserves most feature-relevant information. Whether that information is present in VLAs is what we study.

This \textcolor{Paired2}{Transformer architecture} is then embedded into a \textcolor{Paired4}{VLA model}, where the action $a \in \mathcal{A} \subseteq \mathbb{R}^m$ is computed using the internal representations, i.e., $a=\phi(X_1,\dots,X_L)$.
This map $\phi$ depends on the choice of architecture: for the transformer-based architecture $\phi$ depends only on the final layer representation, $X_L$, while for the transformer-flow-matching hybrid architecture $\phi$ includes the flow matching process that can be conditioned on intermediate layer representations $X_1,\ldots,X_L$.
The pooled vectors $x_\ell$ serve as summaries for our analysis; we do not assume that they alone determine the action.

When deployed in the real world, each sampled action $a$ is executed in the environment, producing a new input and, over repeated interaction, a closed-loop trajectory $\tau \in \mathcal{T} := (\mathcal{W} \times \mathcal{A})^*$ of the form $\tau = (\tau_0, \tau_1, \dots)$, where $\tau_t := (w_t, a_t)$ is the $t$-th transition.
The $(\cdot)^*$ operator captures finite, variable-length trajectories.

\subsection{Feature Observability and Controllability}\label{sec:definitions}

A \emph{feature} is a measurable attribute of the model's output or the robot's behavior during execution, such as an action displacement, grasp modality, or the identity of a manipulated object.
Formally, a feature is a map $\feature : \mathcal{T} \to \mathbb{R}^n$ assigning to each trajectory a value $\feature(\tau)$; this may be a direct readout of the trajectory, but can also be a label provided by a human annotator inspecting the rollout.

In this paper, we explore how the structure of the representation space in VLAs can be leveraged to (i) observe features, and (ii) steer features into a desired region by controlling their representations.
To that end, we introduce the concepts of
\emph{feature-observability} and \emph{feature-controllability}. Both are properties of a feature relative to a collection of inputs to the VLA: the requirement is that a \emph{single} map works for every input in that collection, so the property is a statement
about the representation space at a layer rather than about an individual rollout.

\begin{definition}[Feature-Observability]
    A feature $\feature$ is \emph{observable at layer} $\ell$ if there exists a single map $f_\ell\in\mathcal F$ such that $f_\ell(x_\ell)=\feature(\tau)$ for every input.
    The map $f_\ell$ is referred to as the \emph{observer}.
\end{definition}

\begin{definition}[Feature-Controllability]
    Given a target set $\mathcal C\subseteq\mathbb R^n$, a feature $\feature$ is \emph{controllable to $\mathcal C$ at layer} $\ell$ if there exists a single map $g_\ell \in\mathcal G$ such that, for every input, setting $X_{\ell} = g_\ell(\mathsf{T}_\ell(X_{\ell-1}))$ yields a trajectory $\tilde\tau$ with $\feature(\tilde\tau)\in\mathcal C$. The map $g_\ell$ is referred to as the
    \emph{controller}.
\end{definition}

\begin{remark}\label{remark:feature_scope}
    The definitions above admit a broad range of features. One can distinguish (a) \underline{output-level features}, assessed at a specific point in the trajectory, $\feature(\tau_t)$; and (b) \underline{trajectory-level features}, assessed on the full trajectory, $\feature(\tau)$.
    Consequently a feature need not be observable or controllable at every time step: an eventual grasp modality may not yet be distinguishable early in execution, while a choice already realized cannot be undone by later interventions.
\end{remark}

In the general case, the set of observers and controllers considered are $\mathcal F\subseteq\{f:\mathbb R^d\to\mathbb R^n\}$ and $\mathcal G\subseteq\{g:\mathbb R^{N \times d}\to\mathbb R^{N \times d}\}$.
In practice the observer and controller are usually learned rather than prescribed, so $\mathcal F$ and $\mathcal G$ are restricted to a specific structure (linear maps, one-layer MLPs, etc.) and fitted on a collection of inputs drawn from rollouts together with their representations and feature values.

We state both notions exactly, as the idealized versions make the structure clearest.
Empirically neither holds exactly: we report the accuracy with which a fitted observer recovers $\feature(\tau)$, and the rate at which a controller lands in $\mathcal C$, and read these as graded evidence for the corresponding property.

Given the definitions above, the open question is which layers admit observers and controllers for features of interest, and how small the classes $\mathcal F$ and $\mathcal G$ can be while still doing so.
In what follows we propose that linear choices suffice: linear observers recover features of interest from $x_\ell$, and linear, additive controllers steer the resulting behavior toward specified targets $\mathcal C$ during execution.

\section{Approach}\label{sec:approach}

In this section we instantiate the two introduced concepts, \emph{feature-observability} and \emph{feature-controllability}, with concrete choices of $\mathcal F$ and $\mathcal G$.
We propose a linear \emph{observer}, which extracts the
value of an interpretable feature from the model's internal state, and an additive, linear \emph{controller}, which intervenes on that state to steer the resulting behavior toward a target set.
Both are cheap enough to apply online during execution, requiring a single matrix multiply per forward pass.
We evaluate how well they hold empirically in Section~\ref{sec:experiments}.

\subsection{Designing the Feature Observer}\label{sec:feature-observer}

We design the observer map at layer $\ell$, $f_{\ell}(\cdot)$, to be a linear function, i.e. we restrict $\mathcal F$ to be the set of linear maps from $\mathbb R^d$ to $\mathbb R^n$. This is, given a representation $x_{\ell}\in\mathbb R^d$, the feature of interest is predicted as $\hat{\feature} = f_{\ell}(x_{\ell})$, with
\begin{equation}\label{eqn:observer}
    f_{\ell}(x):=W_{\ell}^\top x+b_{\ell},
\end{equation} 
where $W_{\ell}\in\mathbb R^{d\times n}$ and $b_{\ell}\in\mathbb R^{n}$ are learned from data. Specifically, we leverage pairs of inputs ($w$ containing images, prompt and potentially robot states) to a feature of interest (denoted as $\feature\in\mathbb R^n$). Given a set of input-feature pairs $\{w^{(i)},\feature^{(i)}\}_{i=1}^K$, a set of activation-feature pairs $\{x_{\ell}^{(i)},\feature^{(i)}\}_{i=1}^K$ can be collected at each layer $\ell$ by simply propagating each input sequence $w^{(i)}$ up until layer $\ell$ through the transformer architecture, whose output is $x_{\ell}^{(i)}$.
Then, $W_{\ell}$ and $b_{\ell}$ are learned by minimizing a task-appropriate loss over $\{x_{\ell}^{(i)},\feature^{(i)}\}_{i=1}^K$:
\begin{equation}\label{eqn:train_loss}
W_{\ell}, b_{\ell} = \operatorname*{arg\,min}_{W,\,b} \sum_{i=1}^K \mathcal{L}\bigl(W^\top x_\ell^{(i)} + b, \feature^{(i)}\bigr),
\end{equation}
where $\mathcal{L}$ is the mean squared error loss for continuous features ($\feature \in \mathbb{R}^n$) and the cross-entropy loss for discrete features ($\feature \in \{0,1\}^n$).
Algorithm \ref{alg:offline} summarizes the steps to train the observer at layer $\ell$, $f_{\ell}(\cdot)$.

\begin{algorithm}[b]
\caption{Learning linear observer $f_{\ell}$ (Eq. \eqref{eqn:observer}) at layer $\ell$}
\label{alg:offline}
\begin{algorithmic}[1]

\State \textbf{Input:} Labeled dataset $\{(w^{(i)}, \feature^{(i)})\}_{i=1}^K$
\State \textbf{Output:} $W_{\ell}\in\mathbb R^{d\times n}$ and $b_{\ell}\in\mathbb R^n$

\For{$i = 1,\dots,K$} \Comment{\textcolor{gray}{In parallel for all $i=1,\dots,K$}}
\State $x_0^{(i)} \leftarrow E(w^{(i)})$
\Comment{\textcolor{gray}{Compute initial embedding}}

\For{$\ell = 0,\dots,L-1$}\Comment{\textcolor{gray}{Until layer $L$}}
    \State $x_{\ell+1}^{(i)} \leftarrow \text{Eq.~\eqref{eqn:layers}}$
    \Comment{\textcolor{gray}{Get layer $\ell$ representation}}
\EndFor
\EndFor

\State $(W_{\ell}, b_{\ell}) \leftarrow
\text{Eq.~\eqref{eqn:train_loss}}$
\Comment{\textcolor{gray}{Train linear observer at layer $\ell$}}

\State \Return $W_{\ell}, b_{\ell}$

\end{algorithmic}
\end{algorithm}

The choice of a linear map is motivated by the linear separability hypothesis, widely explored in LLMs \cite{park2024linear}. We note that the presented method holds for generalized versions of the linear observer, such as $f_{\ell}(x)=\nu(W_{\ell}^\top x+b_{\ell})$, where $\nu$ is some known monotonic nonlinearity of choice.

\subsection{Designing the Feature Controller}\label{sec:feature-controller}

We design the controller map at layer $\ell$ to act additively on the full token
representation, broadcasting a single perturbation $u_\ell\in\mathbb R^d$ to every token.
That is, we restrict $\mathcal G$ to maps of the form
\begin{equation}\label{eqn:controller}
    g_{\ell}(X) := X+\mathbf 1_N\,u_{\ell}(x)^\top,
    \qquad x=\tfrac1N X^\top\mathbf 1_N,
\end{equation}
where $\mathbf 1_N\in\mathbb R^N$ is the all-ones vector and the perturbation depends on $X$ only through its mean-pooled summary $x$. This shifts the pooled representation by exactly $u_\ell:=u_\ell(x_\ell)$.

Broadcasting is not an arbitrary choice: it is the least-norm way to realize a prescribed shift of the pooled summary.

\begin{proposition}[Minimum-norm realization of a pooled perturbation]
\label{prop:minimal_broadcast}
For a prescribed pooled perturbation $u\in\mathbb R^d$, the broadcast perturbation $\Delta X^\star=\mathbf 1_Nu^\top$ is the unique minimizer of $\|\Delta X\|_F^2$ over $\Delta X\in\mathbb R^{N\times d}$ satisfying $\frac{1}{N}\Delta X^\top\mathbf 1_N=u$. Its squared Frobenius norm is $N\|u\|_2^2$.
\end{proposition}

\begin{proof}
Write $\delta_j=\Delta X_{j,:}^\top$, so that $\sum_{j=1}^N\delta_j=Nu$.
Expanding about $u$,
\[
\|\Delta X\|_F^2=\sum_{j=1}^N\|\delta_j\|_2^2
=N\|u\|_2^2+2u^\top\!\!\sum_{j=1}^N(\delta_j-u)+\sum_{j=1}^N\|\delta_j-u\|_2^2,
\]
and the middle term vanishes by the constraint. Hence $\|\Delta X\|_F^2\ge N\|u\|_2^2$, with equality iff $\delta_j=u$ for all $j$.
\end{proof}

Following prior work in LLMs \cite{liseco}, we compute $u_{\ell}$ as the solution to the optimization problem:
\begin{subequations}\label{eqn:liseco_optimization}
    \begin{align}
    u_{\ell} = \underset{u\in\mathbb R^d}{\text{argmin}} & \qquad \|u\|_2^2 \label{eqn:global_cost_relx_cont_range}\\
    s.t. 
    &  \qquad f_{\ell}(x_{\ell} + u)\in\mathcal C,
    \label{eqn:constr_probe_relx_range} 
    \end{align}
\end{subequations} 
where $\mathcal C\subseteq \mathbb R^n$ is the desired set of feature values, and $f_{\ell}(\cdot)$ is the observer map at layer $\ell$.
This optimization finds the smallest pooled perturbation that places the observer output in $\mathcal C$, and by Proposition~\ref{prop:minimal_broadcast} it is also the smallest intervention on the full token representation, since $\|\Delta X\|_F^2=N\|u\|_2^2$.

In general, the constraint in \eqref{eqn:constr_probe_relx_range} may be nonconvex, depending on the form of $f_{\ell}$ and the geometry of $\mathcal C$. To obtain a tractable solution, we use the linear observer in \eqref{eqn:observer}, and consider a scalar
feature with target interval
$\mathcal C=[\feature_{\min},\feature_{\max}]\subseteq\mathbb R$,
where $\feature_{\min}\le\feature_{\max}$.
For $W_\ell\ne0$, the solution to
\eqref{eqn:liseco_optimization} has the following closed form
\cite[Theorem~4.1]{liseco}:
\begin{subequations}\label{eq:u_closed_form}
\begin{align}
u_{\ell} &=
(\feature_{\max} - \hat{\feature}_{\ell})\dfrac{W_{\ell}}{\|W_{\ell}\|_2^2},
&& \text{if } \hat{\feature}_{\ell} > \feature_{\max}, \label{eq:u_closed_form:a} \\
u_{\ell} &=
(\feature_{\min} - \hat{\feature}_{\ell})\dfrac{W_{\ell}}{\|W_{\ell}\|_2^2},
&& \text{if } \hat{\feature}_{\ell} < \feature_{\min}, \label{eq:u_closed_form:b} \\
u_{\ell} &= 0,
&& \text{otherwise},
\label{eq:u_closed_form:c}
\end{align}
\end{subequations}
where $\hat{\feature}_{\ell}=f_\ell(x_{\ell})=W_{\ell}^\top x_{\ell} + b_{\ell}$ is the layer $\ell$ observation.

\begin{figure*}[t]
  \centering
  \includegraphics[width=1.02\textwidth]{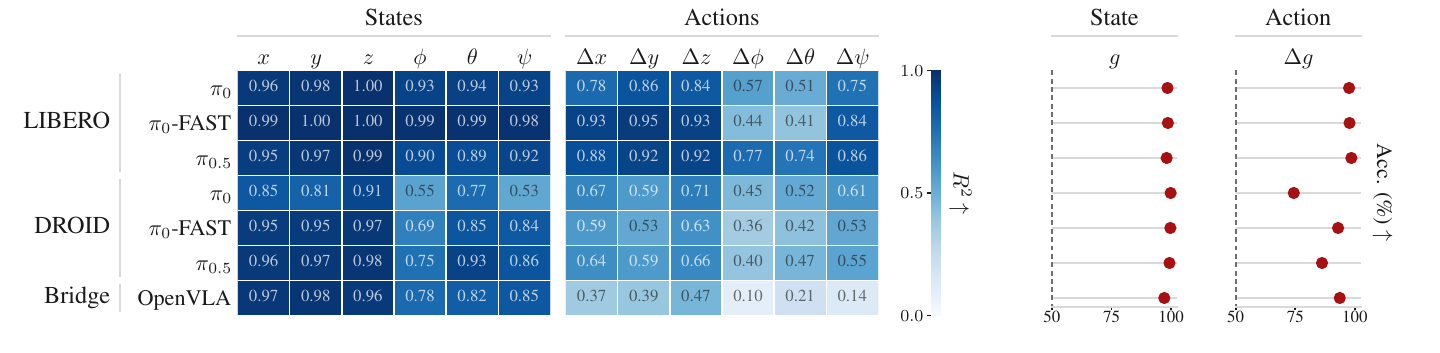}
  \vspace{-2em}
  \caption{Linear observability of robot states and actions inside four vision-language-action models.
  Each row is one model evaluated on one dataset.
  For every quantity a linear probe is trained per layer on pooled Transformer activations, and scored on a held-out 10\% split.
  Heatmaps report $R^2$ for the six continuous states and actions, taking the best of the probed layers per quantity.
  Dot plots report accuracy on binary quantities, gripper state $g$ and gripper command $\Delta g$, where 50\% is chance.
  }
  \label{fig:model_dataset_observability}
  \vspace{-1em}
\end{figure*}

Observability and controllability are independent properties~\cite{kalman1960new}, but methods from the LLM literature show that steering using a learned readout of the feature~\cite{act,liseco} outperform those injecting directions without one~\cite{turner2023activation}.
Therefore, our proposed controller relies on the observer.

\subsection{Observing and Controlling Features in VLAs}\label{sec:inference-time}

We illustrate how to leverage the proposed feature observer and feature controller in practice to align VLA outputs with user preferences. Given
\begin{itemize}
    \item a \emph{feature observer}, $f_{\ell}(\cdot)$, with linear structure as defined in Equation \eqref{eqn:observer}, where linear parameters $W_{\ell}$ and $b_{\ell}$ are computed before inference time via Algorithm \ref{alg:offline}; and
    \item a \emph{feature controller}, $g_{\ell}(\cdot)$, with linear structure as defined in Equation \eqref{eqn:controller}, where the additive $u_{\ell}$ is computed according to Equation \eqref{eq:u_closed_form},
\end{itemize} 
the goal is to integrate them into the inference time computations at layer $\ell$ to achieve precise representation control and downstream alignment of the VLA output.

\begin{algorithm}[b]
\caption{Inference forward-pass in a VLA-embedded Transformer architecture with $\feature$-observer and $\feature$-controller}
\label{alg:inference}
\begin{algorithmic}[1]

\State \textbf{Input:} $w$,  $\Lambda_O$, $\Lambda_C$, $W_{\ell\in\Lambda_O}$, $b_{\ell\in\Lambda_O}$, $\feature_{\min}$, $\feature_{\max}$
\State \textbf{Output:} $x_1,\dots,x_L$
\State $x_0 \leftarrow E(w)$
\Comment{\textcolor{gray}{Compute initial embedding}}
\For{$\ell = 1,\dots,L$} \Comment{\textcolor{gray}{\textbf{Forward-pass through layers}}}
\State $x_{\ell} \leftarrow \text{Eq.~\eqref{eqn:layers}}$
    \Comment{\textcolor{gray}{Compute layer $\ell$ representation}}
\If{$\ell \in \Lambda_O$}\Comment{\textcolor{gray}{\textbf{Feature-observability}}}
\State $\feature_{\ell} \leftarrow \text{Eq.~\eqref{eqn:observer}}$
\Comment{\textcolor{gray}{Compute observation}}
\If{$\ell \in \Lambda_C$}\Comment{\textcolor{gray}{\textbf{Feature-controllability}}}
\State $u_{\ell} \leftarrow \text{Eq.~\eqref{eq:u_closed_form}}$
    \Comment{\textcolor{gray}{Compute control input}}
\State $x_{\ell} \leftarrow \text{Eq.~\eqref{eqn:controller}}$ \Comment{\textcolor{gray}{Apply intervention}}
\EndIf
\EndIf
\EndFor
\State \Return $x_1,\dots,x_L$

\end{algorithmic}
\end{algorithm}

Let us consider two set of layers: (i) $\Lambda_O\subseteq [0,\dots,L]$ where feature-observability is desired, and (ii) $\Lambda_C\subseteq [0,\dots,L]$ where feature-controllability is desired. We note that, given that the proposed controller requires observations, $\Lambda_C \subseteq \Lambda_O$.
The inference-time forward pass equipped with observer and controller for feature $\feature$ is described by Algorithm \ref{alg:inference}.

The forward-pass presented in Algorithm \ref{alg:inference} only differs from a standard un-observed and un-intervened Transformer forward-pass in steps 6-12.
A crucial advantage of choosing a linear observer and a linear closed-form controller is that computations in step 7, 8 and 9 introduce minimal overhead, resulting in a negligible increase in extra runtime.

\section{Experiments}\label{sec:experiments}

In this section, we present the experiments that explore the concepts of feature-observability and feature-controllability in frontier VLA policies: the transformer-based OpenVLA and $\pi_0$-FAST and the transformer-flow-matching hybrid models $\pi_0$ and $\pi_{0.5}$.
We first compare how accurately robot states and actions can be observed in representation space using a linear observer across these architectures.
Then, we show that targeted perturbations in the representation space smoothly affect VLA outputs and that precise representation control is achieved.
Lastly, we study how minimal control interventions to the representation space lead to effective VLA steering in \emph{closed-loop} deployment on hardware.

\subsection{Observing Features in VLA Representations}\label{sec:feature_obs}

We assess whether robot-relevant features can be predicted using the linear observer defined in Equation~\eqref{eqn:observer}, and how prediction performance varies across models and datasets.
We evaluate $\pi_0$, $\pi_0$-FAST, and $\pi_{0.5}$ on LIBERO~\cite{liu2023libero} and DROID~\cite{khazatsky2024droid}, and OpenVLA~\cite{OpenVLA} on BridgeData V2~\cite{walke2023bridgedata}.
We focus on robot states and actions, which are available in these datasets, and fundamental robot quantities.
Following Remark~\ref{remark:feature_scope}, these are output-level features, while trajectory-level features are evaluated in Section~\ref{sec:closed_loop}.

The state features are $s=(x,y,z,\phi,\theta,\psi,g)$, comprising Cartesian end-effector position, roll, pitch, yaw, and normalized gripper aperture.
Action features comprise translational and rotational commands $(\Delta x,\Delta y,\Delta z,\Delta\phi,\Delta\theta,\Delta\psi)$ and the gripper command $\Delta g$, following each dataset's conventions.
For chunked actions, labels are averaged over the chunk; OpenVLA uses single-step action labels.
Continuous quantities are predicted using regression probes, while gripper state and command are predicted using binary classifiers.

For each model--dataset pair, we construct input--feature pairs $\{(w^{(i)},\feature^{(i)})\}_{i=1}^K$, where, for example, $\feature=\Delta x\in\mathbb R$.
Inputs contain scene images, task instructions, and proprioceptive measurements as supported by each model; OpenVLA receives a single image and no proprioceptive measurements.
We pass these inputs through the frozen models and collect pooled representations at each layer, yielding activation--feature pairs $\{(x_\ell^{(i)},\feature^{(i)})\}_{i=1}^K$.
For all models, we probe the VLM prefix, excluding action-expert activations in hybrid architectures.
For autoregressive models, activations are collected during the first forward pass, before action-token generation.
Observers are trained separately for each feature and layer using Algorithm~\ref{alg:offline}, and evaluated on a held-out $10\%$ split.
Figure~\ref{fig:model_dataset_observability} reports the highest score across layers.

\begin{figure}[t]
    \centering
    \includegraphics[width=0.9\columnwidth]{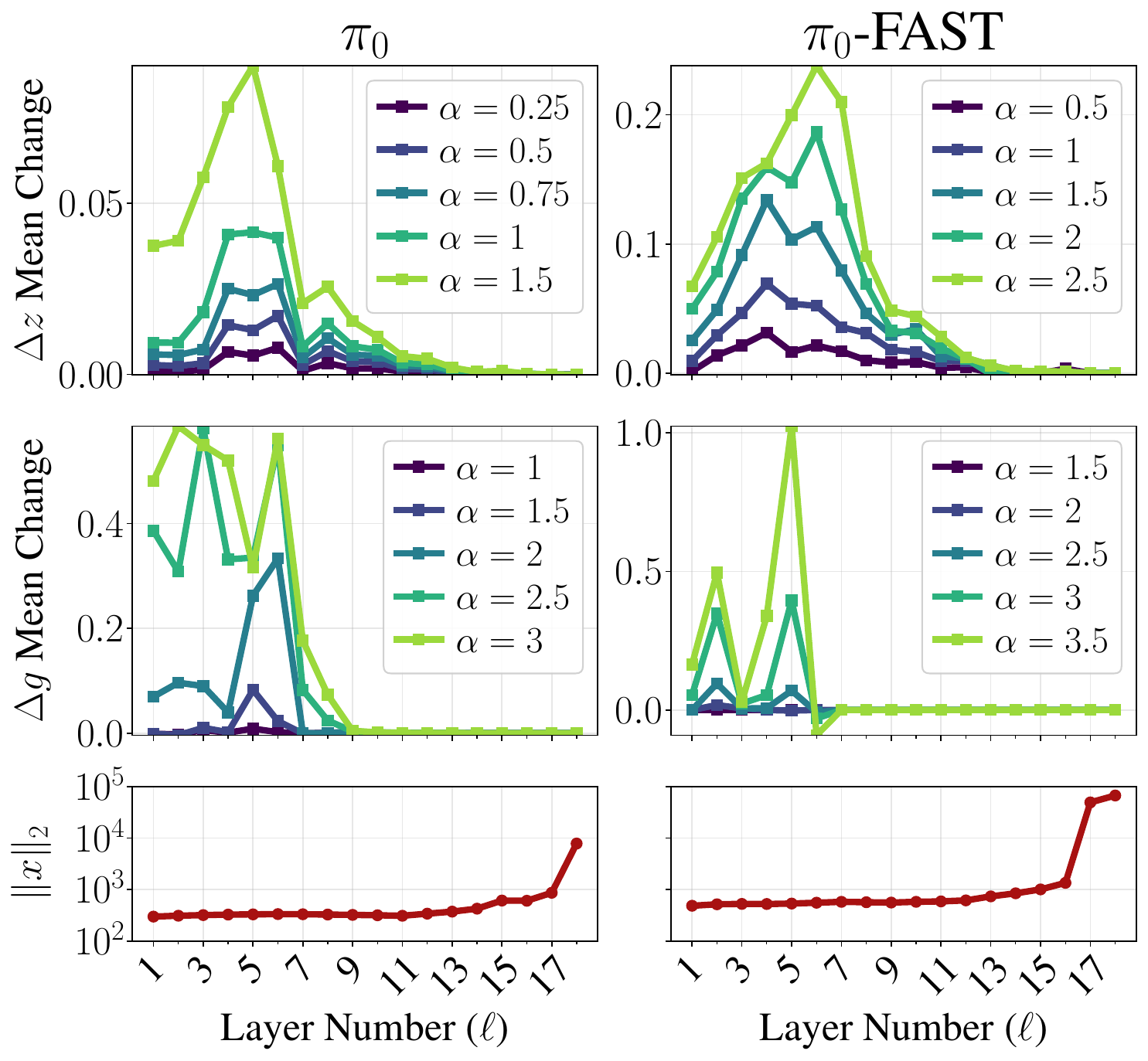}
    \vspace{-0.5em}
    \caption{Output sensitivity to observer-aligned representation interventions in the LIBERO checkpoints of $\pi_0$ (left) and $\pi_0$-FAST (right). Top and middle: mean changes in vertical displacement $\Delta z$ and gripper command $\Delta g$ relative to the unmodified model, as a function of intervention layer, for different strengths $\alpha$.
    Bottom: representation $L_2$-norm across layers, on a logarithmic scale.
    }
    \vspace{-1em}
    \label{fig2}
\end{figure}

Figure~\ref{fig:model_dataset_observability} shows that all evaluated model--dataset pairs
contain linearly predictive information about robot states and actions.
State features are recovered more accurately than action features throughout: Cartesian
state probes achieve $R^2=0.81$--$1.00$ and orientation probes $R^2=0.53$--$0.99$, while
the corresponding action probes fall well below this.
Similarly, gripper-state classification is close to perfect, while gripper-command
accuracy shows a larger spread.
We also note that $\pi_{0.5}$ on average has the strongest decodability of continuous features, but not for binary (gripper) features.

Action decodability is markedly lower on the real-world datasets while state decodability
is not, dropping by roughly a third from LIBERO to DROID with state scores essentially
unchanged. Since the gap is specific to actions, one explanation is that the LIBERO
policies encode a comparatively rigid state-to-action mapping fit to scripted
demonstrations, whereas policies trained on diverse teleoperation data retain more distributed and multimodal action representations.
Real-world action targets are also intrinsically noisier, so lower $R^2$ need not by itself imply less linear structure.

Overall, linear predictability of state and action features is present in autoregressive
and transformer--flow-matching architectures alike, with the simulated--real gap in action
decodability larger than any gap between architectures.

\newpage
\subsection{Controlling Features Through Representation Interventions}\label{sec:feature_ctrl}

Feature predictability alone does not establish that a representation can be used to steer model behavior. We therefore examine two requirements for representation control: whether observer-aligned perturbations produce graded changes in the generated actions, and whether the controller can place individual observer predictions within a prescribed target.

\textbf{Output sensitivity to representation interventions.}
We compare the LIBERO checkpoints of $\pi_0$ and $\pi_0$-FAST to examine flow-matching and autoregressive action generation within the same model family. For each scalar action feature, we perturb the representation along its linear observer direction,
$x_\ell \leftarrow x_\ell + \alpha W_\ell$,
where $\alpha$ controls intervention strength. Only one layer is perturbed per generation.
We sweep the layer and intervention strength, measuring the mean change in the corresponding output feature relative to the unmodified model across \num{50} LIBERO episodes.

Across the six continuous action dimensions, increasing $\alpha$ produces graded, generally ordered changes in the mean output over the tested intervention strengths. Figure~\ref{fig2} illustrates this behavior for the vertical displacement command $\Delta z$. The trend appears in both models, indicating that observer-aligned interventions can modulate continuous actions under both generation mechanisms. Similar continuous-action responses are observed for the LIBERO and DROID checkpoints of $\pi_{0.5}$, which we use in subsequent closed-loop experiments.
The binary gripper response is less regular, while still showing an ordering with increasing $\alpha$.
Lower sensitivity in later layers may partly be attributed to increasing representation scales, which can reduce the relative effect of a fixed perturbation.

\begin{figure}[t]
    \centering
    \includegraphics[width=\columnwidth]{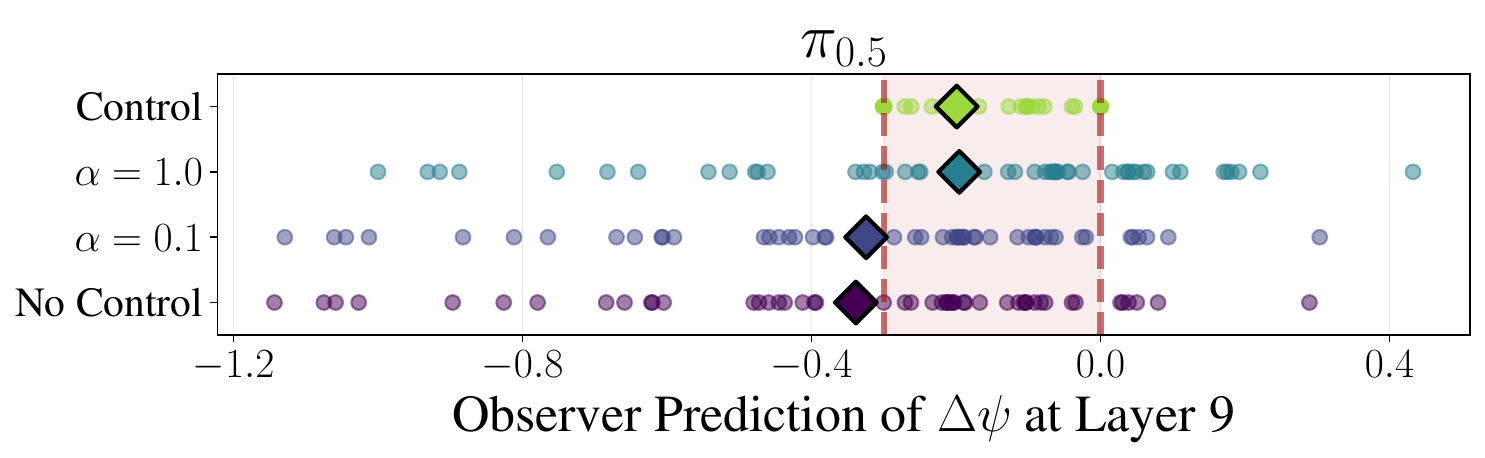}
    \vspace{-1.5em}
    \caption{Representations projected into the classifier image space at layer 9 of $\pi_{0.5}$, under different interventions. Individual samples are shown with means indicated by solid diamonds. The controller constrains the representations to lie within the target bounds $[\feature_{\min}, \feature_{\max}]$, while baseline interventions and un-intervened representations fall outside.}
    \label{fig:representation_control}
    \vspace{-1em}
\end{figure}

\textbf{Control of observer predictions.}
We next evaluate the minimal controller in Equation~\eqref{eq:u_closed_form} in the output space of the scalar affine observer from Equation~\eqref{eqn:observer}. A fixed additive perturbation shifts all observer predictions by the same amount, preserving their spread.
Consequently, it can move the mean prediction into the target interval while leaving individual samples outside, as Figure~\ref{fig:representation_control} illustrates.
The proposed controller instead adapts the intervention to each representation, satisfying $f_\ell(\widetilde x_\ell)
    \in [\feature_{\min},\feature_{\max}]$
by construction of the constraint in Equation~\eqref{eqn:liseco_optimization}.
Figure~\ref{fig:representation_control} demonstrates this empirically.
This guarantee concerns the observer output; behavioral constraint satisfaction during closed-loop execution is evaluated in the next section.

\newpage
\subsection{Observing and Controlling Features in Closed-Loop}\label{sec:closed_loop}

\begin{figure}[t]
    \centering
    \includegraphics[width=\columnwidth]{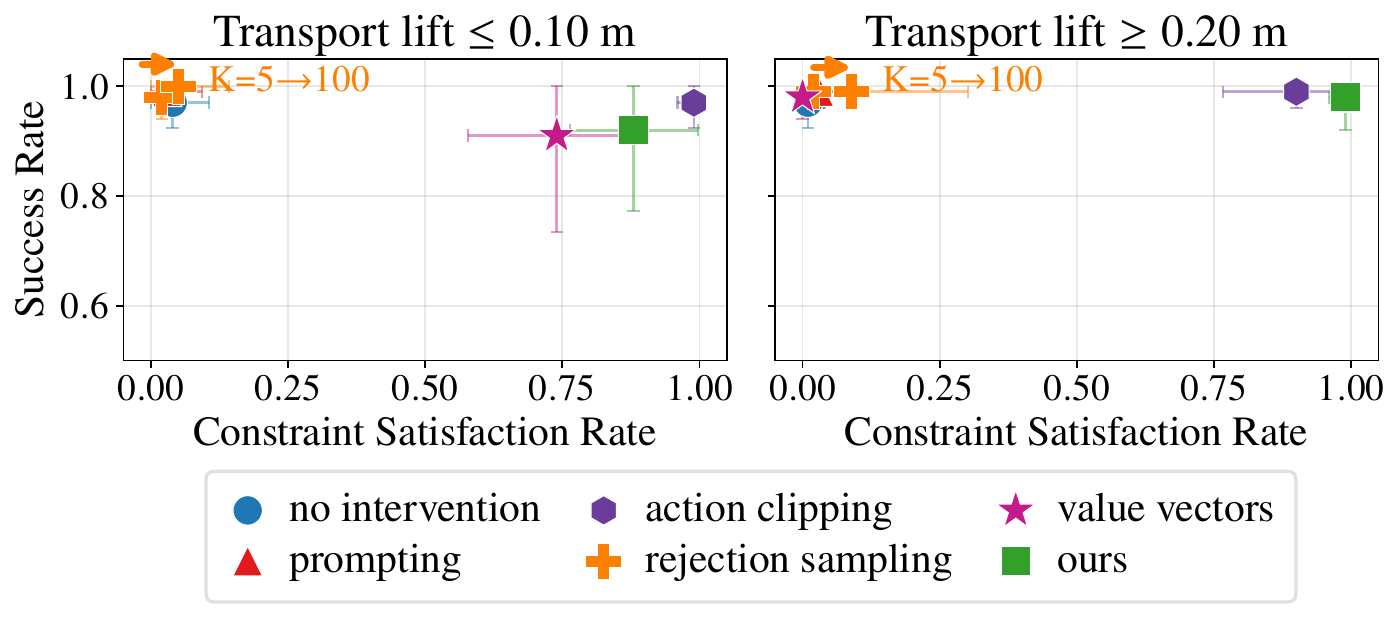}
    \vspace{-1.5em}
    \caption{Task success rate vs. constraint satisfaction in LIBERO-Spatial across 100 rollouts per method (10 tasks × 10). The lift during the transport phase is constrained above/below. Whiskers show $\pm 1$ standard deviation across tasks.}\label{fig:lift_constraint}
    \vspace{-1em}
\end{figure}

The preceding experiments examined state and action observability and the effects of representation interventions on model outputs. We now evaluate whether these interventions steer executed behavior during closed-loop interaction, including grasp modalities and object choices, while preserving task performance. We use frozen $\pi_{0.5}$ checkpoints~\cite{intelligence2025pi05} in LIBERO simulation~\cite{liu2023libero} and on the DROID hardware platform~\cite{khazatsky2024droid}, with policy inference on a single NVIDIA RTX 5090 GPU. 

\textbf{Evaluation and baselines.}
We consider transport-lift constraints, preferred grasp modalities, and task-specific object-selection constraints.
The three simulation experiments comprise \num{100} rollouts each, while each of the three hardware tasks uses \num{50} rollouts per method.
Constraint satisfaction is the fraction of rollouts for which $\feature(\tau)\in\mathcal C$; for categorical features, this is the frequency of the preferred grasp or designated safe object.
Task success is evaluated separately.

The main baselines are the unmodified policy, prompt interventions, and value-vector steering~\cite{haon2025mechanistic}.
For lift constraints, we additionally evaluate two output-side baselines: rejection sampling and action clipping. Rejection sampling draws up to $K$ action chunks at each replan, selecting a satisfying candidate when available and otherwise executing the least-violating candidate. 
Action clipping directly modifies the commanded motion, using a safety margin to account for discrepancies between commanded and executed trajectories.
The \num{50} base policy rollouts are re-used to fit the observers.

\textbf{Constraining transport motion.}
We impose either an upper bound of $0.10$\,m or a lower bound of $0.20$\,m on the vertical lift during the task's transport phase. Constraints apply over this phase, and task success requires completion of the original LIBERO task. Figure~\ref{fig:lift_constraint} shows that representation control substantially increases constraint satisfaction while retaining high task success.
Prompting, on the other hand, fails to meaningfully alter the policy's behavior.
Increasing the rejection-sampling budget $K=5$ to $K=100$ leaves satisfaction below \num{10}\% for both bounds, while value vectors achieve approximately \num{74}\% for the upper bound and \num{0}\% for the lower bound.
As expected for these geometric constraints, action clipping performs well. In our experiments, only the lower-bound constraint requires tuning: the limited vertical displacement achievable in a single control step necessitates a margin (of $4$ cm) that activates corrective motion early enough to reach the required lift during transport.

Rejection sampling also incurs substantial computational overhead. Even with shared prefix computation and batched action-expert evaluations, median inference time increases by \num{16}\% at $K=5$ and \num{583}\% at $K=100$, compared with approximately \num{1}\% for our closed-form representation interventions (measured over \num{50} forward passes).

\begin{figure}[t]
    \centering
    \includegraphics[width=\columnwidth]{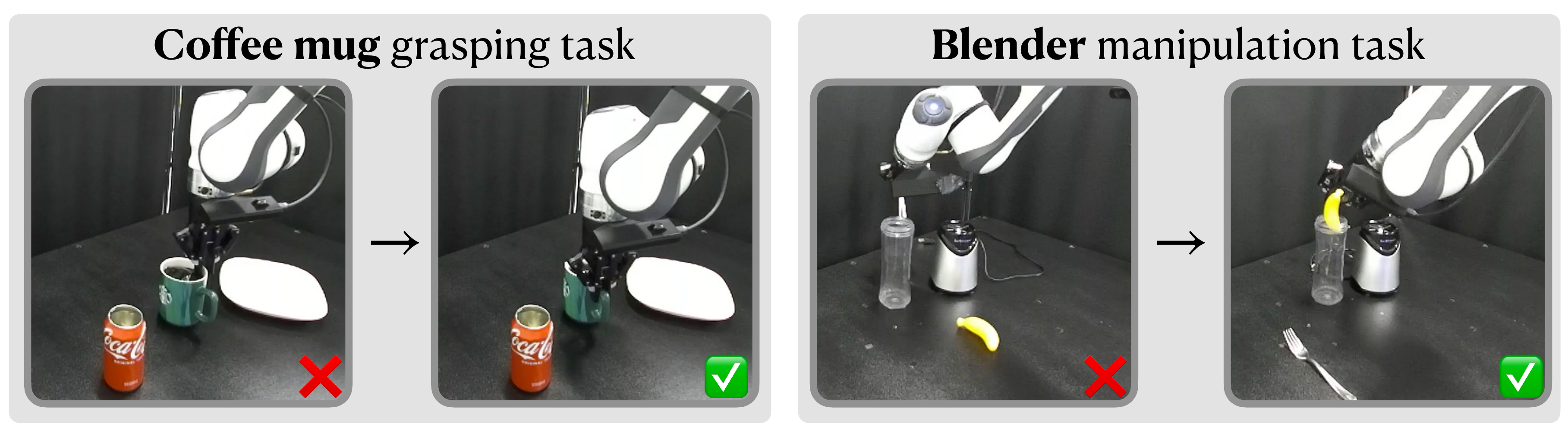}
    \vspace{-1.75em}
    \caption{Examples of closed-loop hardware behaviors. Left: grasping a coffee-filled mug by the rim (undesired) or handle (preferred). Right: selecting a metal fork (unsafe) or banana (safe) for a blender. Arrows indicate desired behavior change.}\label{fig:hardware_tasks}
    \vspace{-1em}
\end{figure}

\textbf{Aligning grasp modalities with user preferences.}
We next constrain how the robot grasps an object. In LIBERO, the task is to pick up a wine bottle, with a more stable, preferred grasp at the base rather than the top.
As shown in Table~\ref{tab:grasping}, the base policy selects the preferred modality in only \num{28}\% of rollouts.
The better of two prompts emphasizing grasp location or approach direction raises this to \num{40}\%, while value-vector steering reduces it to \num{3}\%.
Representation control makes the preferred modality dominant at \num{73}\%.

On hardware, the robot must pick up a coffee-filled mug by its handle; a rim grasp places a gripper finger in the liquid. The base policy grasps the handle in only \num{14}\% of rollouts. Prompting and value vectors increase this to \num{46}\% and \num{38}\%, respectively, while our method reaches \num{74}\%. All methods successfully grasp the mug, so the improvement concerns the executed grasp modality rather than grasp success. Figure~\ref{fig:hardware_tasks} illustrates the preferred and undesired behaviors.
In both tasks, representation control makes the preferred modality dominant, although some rollouts still violate the preference.

\begin{table}[b]
  \vspace{-1em}
  \centering
  \caption{Grasp modality distribution across \num{100} sampled initial conditions in the LIBERO-Goal scene (wine bottle) and across \num{50} rollouts on hardware (coffee mug).}
  \label{tab:grasping}
  \vspace{-0.5em}
  \begin{tabular}{lcccc}
    \toprule
    & \multicolumn{2}{c}{Wine bottle} & \multicolumn{2}{c}{Coffee mug} \\
    \cmidrule(lr){2-3} \cmidrule(lr){4-5}
    & Base $\uparrow$ & Top $\downarrow$ & Handle $\uparrow$ & Rim $\downarrow$ \\
    \midrule
    Base policy ($\pi_{0.5}$)        & \num{28}\% & \num{72}\% & \num{14}\% & \num{86}\% \\
    Prompting                         & \num{40}\% & \num{60}\% & \num{46}\% & \num{54}\% \\
    Value vectors~\cite{haon2025mechanistic} & \num{3}\% & \num{97}\% & \num{38}\% & \num{62}\% \\
    Ours                              & \bfseries\num[detect-weight=true]{73}\% & \bfseries\num[detect-weight=true]{27}\% & \bfseries\num[detect-weight=true]{74}\% & \bfseries\num[detect-weight=true]{26}\% \\
    \bottomrule
  \end{tabular}
\end{table}

\textbf{Enforcing task-specific semantic constraints.}
Finally, we evaluate object selection in two hardware tasks: choosing a banana rather than a metal fork for a blender, and a glass rather than a plastic bowl for an oven. These constraints concern the identity of the manipulated object. The base policy selects the designated safe object in \num{52}\% and \num{48}\% of rollouts, respectively (Table~\ref{tab:semantic_safety}). Prompting reaches \num{40}\% and \num{62}\%, while value vectors reach \num{64}\% and \num{94}\%. Our method selects the designated safe object in all \num{50} evaluated rollouts of each task. Here, task success means successfully manipulating the selected object, which all methods achieve; full completion of the blender or oven task is not evaluated.
We emphasize that our method does not guarantee safety in these scenarios, and hence can be complementary to output-level filters with action-space guarantees.

\begin{table}[t]
  \centering
  \caption{Object manipulation distribution across \num{50} rollouts for two semantic safety experiments. In each, the safe object should be selected and the unsafe object should be avoided.}
  \label{tab:semantic_safety}
  \vspace{-0.5em}
  \begin{tabular}{lcccc}
    \toprule
    & \multicolumn{2}{c}{Blender} & \multicolumn{2}{c}{Oven (bowls)} \\
    \cmidrule(lr){2-3} \cmidrule(lr){4-5}
    & Banana $\uparrow$ & Fork $\downarrow$ & Glass $\uparrow$ & Plastic $\downarrow$ \\
    \midrule
    Base policy ($\pi_{0.5}$)        & \num{52}\% & \num{48}\% & \num{48}\% & \num{52}\% \\
    Prompting                         & \num{40}\% & \num{60}\% & \num{62}\% & \num{38}\% \\
    Value vectors~\cite{haon2025mechanistic} & \num{64}\% & \num{36}\% & \num{94}\% & \num{6}\% \\
    Ours                              & \bfseries\num[detect-weight=true]{100}\% & \bfseries\num[detect-weight=true]{0}\% & \bfseries\num[detect-weight=true]{100}\% & \bfseries\num[detect-weight=true]{0}\% \\
    \bottomrule
  \end{tabular}
  \vspace{-1em}
\end{table}

\section{Conclusion and Outlook}

This paper introduced a principled framework for understanding and controlling Vision-Language-Action models through their internal representations. We used this framework to study how architectural design choices affect the interpretability and steerability of VLA internal representations, and leveraged these insights to enable reliable inference-time steering.
We found linearly predictive state and action information in both autoregressive and transformer--flow-matching VLAs, with performance varying across models, datasets, and features.
Our lightweight observers and controllers enable behavioral steering without retraining, adding approximately $1\%$ inference overhead.
Closed-loop experiments in the LIBERO simulator and on DROID hardware demonstrate improved constraint satisfaction while retaining task performance.
On hardware, preferred grasps increased by \num{60} percentage points, and the designated safe objects were selected in all $50$ rollouts, beating both prompting and value vector baselines.

\textbf{Limitations and Future Work.}
While our framework successfully demonstrates feature observability and controllability, several directions warrant further investigation.
First, our current approach requires labeled data to train linear observers, which may not always be available in large-scale robotics datasets.
Second, we focused primarily on the Transformer components of VLA architectures; extending our framework to the diffusion or flow-matching heads would enable end-to-end interpretability and control across hybrid architectures.
Finally, establishing principled safety guarantees and bounds on the effects of representation-space interventions would be crucial for deploying these techniques in safety-critical robotic applications.








\renewcommand*{\bibfont}{\footnotesize}
\printbibliography


\end{document}